\pdfoutput=1
\documentclass[11pt]{article}
\usepackage[preprint]{acl}
\usepackage{times}
\usepackage{latexsym}
\usepackage{microtype}
\usepackage{inconsolata}
\usepackage{graphicx}
\usepackage{booktabs}
\usepackage{multirow}
\usepackage{amsmath}
\usepackage{amssymb}
\usepackage{xspace}
\usepackage{subcaption}

\NewDocumentCommand{\whj}{ mO{} }{\textcolor{red}
{\textsuperscript{\textit{Huaijie}}\textsf{\textbf{\small[#1]}}}}
\NewDocumentCommand{\shibo}{ mO{} }{\textcolor{cyan}
{\textsuperscript{\textit{Shibo}}\textsf{\textbf{\small[#1]}}}}
\NewDocumentCommand{\hanze}{ mO{} }{\textcolor{orange}
{\textsuperscript{\textit{Hanze}}\textsf{\textbf{\small[#1]}}}}
\NewDocumentCommand{\shenao}{ mO{} }{\textcolor{blue}
{\textsuperscript{\textit{Shenao}}\textsf{\textbf{\small[#1]}}}}

\newcommand{\ours}{\textsc{OREO}\xspace}

\title{Offline Reinforcement Learning for LLM Multi-Step Reasoning}

\author{Huaijie Wang$^{\diamondsuit*}$, Shibo Hao$^{\clubsuit}\thanks{Equal contribution. $\dagger$ corresponding author}$, Hanze Dong$^{\heartsuit}$, Shenao Zhang$^{\spadesuit}$\\
{\bf Yilin Bao$^{\clubsuit}$,~~
Ziran Yang$^{\clubsuit}$,~~
Yi Wu$^{\diamondsuit\dagger}$}\\
$^{\clubsuit}$UC San Diego,~~ $^{\diamondsuit}$Tsinghua University,\\ $^{\heartsuit}$Salesforce Research,~~$^{\spadesuit}$Northwestern University\\
}
\newtheorem{theorem}{Theorem}

\begin{document}
\maketitle
\begin{abstract}

Improving the multi-step reasoning ability of large language models (LLMs) with offline reinforcement learning (RL) is essential for quickly adapting them to complex tasks. While Direct Preference Optimization (DPO) has shown promise in aligning LLMs with human preferences, it is less suitable for multi-step reasoning tasks because (1) DPO relies on paired preference data, which is not readily available for multi-step reasoning tasks, and (2) it treats all tokens uniformly, making it ineffective for credit assignment in multi-step reasoning tasks, which often come with sparse reward.
In this work, we propose \ours (\underline{O}ffline \underline{RE}asoning \underline{O}ptimization), an offline RL method for enhancing LLM multi-step reasoning. Building on insights from previous works of maximum entropy reinforcement learning, it jointly learns a policy model and value function by optimizing the soft Bellman Equation. We show in principle that it reduces the need to collect pairwise data and enables better credit assignment. Empirically, \ours surpasses existing offline learning methods on multi-step reasoning benchmarks, including mathematical reasoning tasks (GSM8K, MATH), and embodied agent control (ALFWorld). The approach can be extended to a multi-iteration framework when additional resources are available. Furthermore, the learned value function can be leveraged to guide the tree search for free, which can further boost the performance during test time.\footnote{Code available at: \url{https://github.com/jwhj/OREO}}

\end{abstract}

\section{Introduction}

Large Language Models (LLMs) are increasingly applied to complex tasks requiring multi-step reasoning, such as mathematical problem solving~\citep{uesato2022solving, shao2024deepseekmath, hendrycks2021measuring}, embodied agent control~\citep{wang2023voyager, huang2022language, shridhar2020alfworld, xiang2024language}, and web navigation~\citep{deng2024mind2web, zhou2023webarena, koh2024visualwebarena}. Enhancing LLM reasoning with reinforcement learning (RL) has gained significant interest, as it offers the potential for self-improvement and learning without relying on human-labeled trajectories. However, many popular RL algorithms require costly online data collection, either by generating language on-the-fly or interacting with an environment. For instance, tuning LLMs with Proximal Policy Optimization~\citep[PPO,][]{schulman2017proximal} is often prohibitively expensive for most users, which limits practical applications~\citep{hu2023aligning}.

In contrast, offline RL methods, such as Direct Preference Optimization~\citep[DPO,][]{rafailov2024direct}, provide a more practical approach for aligning LLMs with human preferences. These methods enable practitioners to tune models using pre-existing datasets, eliminating the need for live interaction or data generation. However, attempts to enhance LLMs’ multi-step reasoning abilities with DPO may deliver results close to or even worse than simpler methods like SFT~\citep{yuan2024advancing, chen2024noise}. 
% However, attempts to enhance LLMs’ multi-step reasoning abilities with DPO have yet to deliver satisfactory results~\citep{yuan2024advancing, chen2024noise}.
% Additionally, the requirement to collect pairwise preference data for reasoning tasks introduces further complexity.
Additionally, DPO requires pairwise preference data. In multi-step reasoning tasks, however, data normally consists of independent trajectories with sparse rewards indicating success or failure.
A common alternative is to extract correct trajectories from offline datasets and use them for supervised fine-tuning~\citep{zelikman2022star, aksitov2023rest, dong2023raft, paulus2024advprompter}. While this approach is simple and often effective, it fails to fully exploit the offline dataset's potential—particularly the opportunity to learn from failure experience and enhance model robustness~\citep{kumar2022should}.

In this paper, we introduce \ours (\underline{O}ffline \underline{RE}asoning \underline{O}ptimization), an offline RL algorithm designed to enhance LLMs' multi-step reasoning capabilities. Building on insights from the extensive literature on maximum entropy RL~\citep{ziebart2010modeling, nachum2017bridging, haarnoja2017reinforcement}, especially Path Consistency Learning~\citep{nachum2017bridging}, \ours jointly learns a policy model and a value function by optimizing the soft Bellman Equation. 
% \iffalse
% Upon revisiting DPO within our framework, we demonstrate that two unreasonable assumptions for multi-step reasoning in DPO have been eliminated: the reliance on the Bradley-Terry (BT) preference model and the relaxation of the soft Bellman Equation from every time step to the whole trajectory (See more details in Section~\ref{sec:dpo})
% %$~\whj{We haven't introduced BT model \& soft Bellman Equation yet. We should focus on the main idea, i.e., pairwise data \& credit assignment}.
% % \shibo{I feel adding more details can make the introduction richer.. Otherwise it's too dry. Also the connection is one of our contribution. There are also other works removing the need of pairwise data.}
% Consequently, \ours does not require pairwise preference data collection and enables finer-grained credit assignment, which is especially critical as the correctness of reasoning trajectories often depends on a few key tokens.
% \fi
\ours can leverage unpaired data with only sparse rewards and enables finer-grained credit assignment, which is especially critical as the correctness of reasoning trajectories often depends on a few key tokens. Additionally, \ours can be extended into an iterative framework for online exploration. The trained value function is also directly available to guide the step-level beam search at inference time, further boosting performance.

% Empirically, we show that our approach outperforms all baselines including rejection sampling, DPO, and KTO. If new data can be collected from the trained model, \ours can leverage these data to improve model performance further. The value function provided by our method can be applied to guide beam search or choose the best-of-$K$ action. In addition, we compare several variants of the \ours objective and show that the token-level objective achieves both implementation simplicity and performance.
% \shibo{a paragraph of experiment results}
We demonstrate the effectiveness of our approach on both math reasoning (GSM8K, MATH) and embodied agent control (ALFWorld) tasks. It consistently outperforms baseline methods, including rejection sampling, DPO, and KTO, across different model sizes. Notably, we train a 1.5B model to achieve a 52.5\% accuracy on the MATH dataset using only the original training set. Moreover, iterative \ours steadily improves model performance with additional training rounds, whereas baseline methods like rejection sampling exhibit signs of saturation. The value function learned by our method proves highly effective in guiding beam search for math reasoning tasks or selecting the best-of-$K$ actions in embodied agent control. This results in up to a 17.9\% relative improvement over greedy decoding on the MATH dataset.
\section{Related Work}

\subsection{Reinforcement Learning for LLM} 
Reinforcement Learning (RL) has become a standard approach in the post-training stage of LLMs. A widely adopted method, known as reinforcement learning from human feedback (RLHF) \cite{ziegler2019fine,ouyang2022training}, is designed to align LLM responses more closely with human preferences.
%Beyond alignment, RL algorithms also play a critical role in enhancing reasoning capabilities, such as math and coding. \whj{It feels weird to mention reasoning here... Can we just say that RL methods are applied to LLM alignment?}
Traditional RL methods, such as Proximal Policy Optimization (PPO) \citep{schulman2017proximal}, have been extensively used in LLM post-training \citep{achiam2023gpt,team2023gemini,dubey2024llama}. Alternative approaches, such as rejection-sampling-based methods \citep{dong2023raft,gulcehre2023reinforced,zelikman2022star,hoffman2024training}, preference-based reinforcement learning (RL) \citep{rafailov2024direct,xiong2024iterative,ethayarajh2024kto}, and REINFORCE-like RL \citep{williams1992simple,shao2024deepseekmath,li2023remax,ahmadian2024back}, have recently gained traction in the LLM literature.

%These methods are computationally less demanding and have demonstrated significant improvements in instruction following. 
\iffalse
Our method draws inspiration from rich literature on maximum entropy RL~\citep{ziebart2010modeling}, especially Soft Q-Learning~\citep{haarnoja2017reinforcement} and Path Consistency Learning~\citep[PCL,][]{nachum2017bridging}. Relevant to our method, \citet{guo2021efficient} applied PCL and learned soft Q value function with language model logits, \citet{richemond2024offline} proposes Direct Reward Optimization (DRO) based on a similar formulation and applies it to single-trajectory RLHF. In our work, we focus on enhancing LLM's multi-step reasoning ability with offline RL. A concurrent work DQO~\citep{liu2024enhancing} also explores a similar algorithm for LLM multi-step reasoning, and we present a more detailed discussion in Appendix~\ref{sec:dqo}.
\fi

Maximum-entropy RL~\citep{ziebart2010modeling,haarnoja2017reinforcement} aims to maximize the weighted sum of the accumulated reward and the policy entropy. Notable algorithms such as path-consistency learning (PCL)~\citep{nachum2017bridging} and soft actor-critic (SAC)~\citep{haarnoja2018softactorcriticoffpolicymaximum} effectively utilize this framework. Recent works~\citep{guo2021efficient,richemond2024offline,liu2024enhancing} revealed a strong connection between maximum-entropy RL and the RLHF objective, indicating a promising direction to fine-tune LLMs with soft Q-learning-based algorithms. Concurrent with our work, \citet{liu2024enhancing} leverages the SAC framework and derives a similar algorithm for LLM multi-step reasoning. Our method differs in the derivation approach, the inclusion of a KL regularization term, and the exploration of several loss variants. Notably, we provide deeper empirical insights by incorporating diverse domains, the iterative training setting, and value-guided tree search. More discussions and empirical comparisons can be found in Appendix~\ref{sec:liu}.

\subsection{LLM Reasoning}
%The strong reasoning capabilities demonstrated by language models have seen success in both solving complex tasks \citep{cobbe2021training,shi2024can} and in agentic frameworks \citep{yao2022react, yao2024tree,liu2023reason, hao2023reasoning, hao2024training}. 
As an emergent ability of model scale, it is shown that LLMs are able to generate intermediate reasoning steps to solve complex problems, known as ``scratchpad" \citep{nye2021show} or chain-of-thoughts \citep{wei2022chain, kojima2022large}. 
Recent efforts have enhanced LLM reasoning through supervised fine-tuning~\cite{yue2023mammoth, yue2024mammoth2, yu2023metamath, luo2023wizardmath, hao2024training}. 
When human-annotated reasoning trajectories are unavailable, rejection-sampling-based methods have proven effective.
Among them, Self-Taught Reasoner (STaR) \citep{zelikman2022star} generates rationales and fine-tunes on those leading to correct answers. \citet{singh2023beyond} further proposes an iterative approach based on expectation maximization. 
% Latest research explores enhancing reasoning with additional latent compute~\citep{zelikman2024quiet, hao2024training}. 
Recently, the application of RL algorithms to improve LLM reasoning has gained increasing interest~\citep{aksitov2023rest,gou2023critic,dong2024rlhf,havrilla2024teaching,shao2024deepseekmath,zhao2024automatic}, but the direct usages of DPO are not all successful~\citep{yuan2024advancing, chen2024noise} and efficient, as people have to specifically collect pairwise preference data~\citep{chen2024step, song2024trial}. Our method addresses these limitations of DPO in reasoning with a principled solution. Another line of work aims to train a Process Reward Model (PRM), to provide finer-grained feedback on RL. It is typically trained with Monte-Carlo rollout~\citep{wang2024math, wang2024multi, luo2024improve, zhang2024entropy}, which is a special case of the value function learned through our method. We show that our value function enables test-time scaling~\citep{hao2023reasoning, snell2024scaling, wu2024inference, brown2024large, yao2024tree, hao2024llm, liu2024don} to further boost the reasoning performance through tree search.
\section{Preliminaries}
\label{sec:prelim}
\subsection{MDP for LLM Reasoning}

We define the Markov Decision Process (MDP) for LLM reasoning. At each time step, a new token is generated as the action $a_t$. The state is represented as a token sequence. For reasoning tasks that don't involve interactions with the environment, $s_t$ records the context for LLMs, i.e., \( \mathbf{s}_t = (x_0, \dots, x_L, y_0, \dots, y_{t-1}) \), where \( (x_0, \dots, x_L) \) is the input prompt and \( (y_0, \dots, y_{t-1}) \) is the sequence of generated tokens up to step \( t-1 \). %The action \( \mathbf{a}_t \) represents the selection of the next token \( y_{t} \), and 
The transition function $f$ for these tasks deterministically updates the state as \( \mathbf{s_{t+1}} = f(\mathbf{s}_t, \mathbf{a}_t ) = \mathbf{s}_t\mid \mathbf{a}_t \), where $\mid$ is concatenation. 

For those tasks requiring interacting with an external environment, like embodied agent control, the state and transition function is slightly different: if $\mathbf a_t$ is the final token of the agent's response (e.g., \textit{``go to desk 1''}), then $\mathbf s_{t+1}=f(\mathbf{s}_{t}, \mathbf{a}_t) = \mathbf s_t\mid \mathbf a_t\mid \textit{next observation}$.

The reward function \( r(\mathbf{s}_t, \mathbf{a}_t) \) is generally defined for every state-action pair to provide feedback throughout the generation process. However, in this work, we focus on the challenging case where the reward is non-zero only at the terminal step \( T \), reflecting the correctness of the reasoning chain, or whether the task is successfully accomplished.

Following the standard setup in Reinforcement Learning with Human Feedback (RLHF) \citep{ouyang2022training, rafailov2024direct}, a KL-regularization term is introduced to encourage the learned policy to remain close to a reference policy while optimizing for rewards. Therefore, the optimal policy $\pi_\theta$ can be described as follows:

\vspace{-15pt}
\begin{equation}
\small
\begin{aligned}
\pi^*=\arg \max_\pi &\mathbb{E}_{\left(\mathbf{s}_0,..., \mathbf{s}_T\right) \sim \rho_\pi}\sum_{t=0}^T  \biggl[r\left(\mathbf{s}_t, \mathbf{a}_t\right) \\
&- \beta \log \frac{\pi\left(\mathbf{a}_t \mid \mathbf{s}_t\right)}{\pi_\text{ref}(\mathbf{a}_t \mid \mathbf{s}_t)}\biggr],
\end{aligned}
\end{equation}
\vspace{-10pt}

where \( \pi_\text{ref} \) is the reference policy, $\rho_\pi$ is the state-action trajectory distribution generated by following policy $\pi$, and \( \beta \) controls the strength of the regularization. Typically, $\pi_\text{ref}$ is a pre-trained LLM followed by supervised fine-tuning. The discount factor $\gamma$ is normally omitted in the RLHF setting.

\subsection{Soft Bellman Equation}
Entropy-regularized reinforcement learning (RL)~\citep{ziebart2010modeling, nachum2017bridging, haarnoja2017reinforcement} augments the standard reward maximization objective with an entropy term to encourage exploration and improve the robustness of the learned policy. This formulation has a strong connection to entropy-regularized RL, as the Kullback-Leibler (KL) divergence between two distributions can be decomposed into a cross-entropy term and an entropy term, i.e., $D_{\mathrm{KL}}(\pi(\cdot|s) \| \pi_\mathrm{ref}(\cdot|s)) = \mathbb{E}_{\pi}[-\log \pi_\mathrm{ref}(a|s)] - \mathbb{E}_{\pi}[-\log \pi(a|s)]$.

Adapting from the well-established theory in entropy-regularized RL to our setting, we first define the value function $V^\pi$ of a policy, which quantifies the expected KL-regularized reward of a policy $\pi$ from any given state:

\vspace{-20pt}

%To adapt well-established theory and algorithms to our setting, we define the soft state value function and present the soft Bellman equation as the foundation of our proposed method.
\begin{equation}
\small
\begin{aligned}
V^\pi&\left(\mathbf{s}_t\right)=\mathbb{E}_{\left(\mathbf{a}_{t,}\mathbf{s}_{t+1},...\right) \sim \rho_\pi}\sum_{l=0}^{T-t} \biggl[r\left(\mathbf{s}_{t+l}, \mathbf{a}_{t+l}\right)\\ &-\beta \log \frac{\pi(\mathbf{a}_{t+l}\mid \mathbf{s}_{t+l})}{\pi_\text{ref}(\mathbf{a}_{t+l}\mid \mathbf{s}_{t+l})}\biggr]
\end{aligned}
\end{equation}

Compared to the value function in standard RL, which only includes expected rewards, the above definition incorporates an additional KL regularization term: $\beta \log \frac{\pi(\mathbf{a}_{t+l}\mid \mathbf{s}_{t+l})}{\pi_\text{ref}(\mathbf{a}_{t+l}\mid \mathbf{s}_{t+l})}$.

\begin{theorem}\label{thm:consistency}
The optimal policy and its value function satisfy the \textbf{soft Bellman Equation}:
\begin{equation}\label{eq:consistency}
\small
V^*(\mathbf{s}_{t}) -  V^*(\mathbf{s}_{t+1}) = r(\mathbf{s}_t, \mathbf{a}_t) - \beta\log \frac{\pi^*(\mathbf{a_t}\mid \mathbf{s}_t)}{\pi_\mathrm{ref}(\mathbf{a_t}\mid \mathbf{s}_t)}
\end{equation}

where $\mathbf{s}_{t+1} = f(\mathbf{s}_t, \mathbf{a}_t)$.

\end{theorem}
Building on \citet{nachum2017bridging,haarnoja2017reinforcement}, we extend their theorem with a lightweight derivation tailored to our setting, with the proof provided in Appendix~\ref{sec:proof}.

This equation characterizes the relationship between the optimal policy and its value function, providing a theoretical basis for our proposed method. When $\beta=0$, the equation degenerates to the Bellman equation in standard RL. Importantly, when the soft Bellman Equation is always satisfied, the policy and the value function are guaranteed to be the optimal ones:

\begin{theorem}\label{thm:pcl}
If a policy \( \pi(\mathbf{a} \mid \mathbf{s}) \) and state value function \( V(\mathbf{s}) \) satisfy the consistency property (\ref{eq:consistency}) for all states \( \mathbf{s} \) and actions \( \mathbf{a} \) (where \( \mathbf{s}' = f(\mathbf{s}, \mathbf{a}) \)), then \( \pi = \pi^* \) and \( V = V^* \).
\end{theorem}

Similarly, the proof is a simple extension to \citet{nachum2017bridging}. 
Based on Theorem~\ref{thm:pcl}, our proposed method \ours aims to learn both a policy model $\pi_\theta$ and a value model $V_\phi$ towards the optimal policy and value function. This is achieved by minimizing the inconsistency of Soft Bellman Consistency. A more formal description of our method is presented in Section~\ref{sec:method}. 

\subsection{Connection to DPO}
\label{sec:dpo}
In this section, we introduce how DPO can be derived from the formulation above with two additional assumptions. This enables us to understand the limitation of DPO on LLM reasoning from the principle, and motivates us to propose the new method. \citet{rafailov2024r} present a related derivation to analyze the properties of DPO.

First, DPO relaxes the requirements of soft Bellman Equation by telescoping time steps:
\vspace{-5pt}
\begin{equation}\label{eq:dpo}
\small
    \sum_{t=0}^{T-1}r(\mathbf{s}_t, \mathbf{a}_t) = V^*\left(\mathbf{s}_0\right)+\sum_{t=0}^{T-1} \beta \log \frac{\pi^*\left(\mathbf{a}_t \mid \mathbf{s}_t\right)}{\pi_{\mathrm{ref}}\left(\mathbf{a}_t \mid \mathbf{s}_t\right)}
\end{equation}

It then introduces the Bradley-Terry preference model~\citep{bradley1952rank}, which assumes that the probability of one response being preferred over another is determined by the normalized relative exponential rewards of the responses:
\begin{equation}
\small
    p^*\left(\tau^w \succeq \tau^l\right)=\frac{\exp \left(r\left(\mathbf{s}_T^w, \mathbf{a}_T^w\right)\right)}{\exp \left(r\left(\mathbf{s}_T^w, \mathbf{a}_T^w\right)\right)+\exp \left(r\left(\mathbf{s}_T^l, \mathbf{a}_T^l\right)\right)} .
\end{equation}

By maximizing the log-likelihood that a winning response is preferred over a losing response with a preference dataset \(D=\{(\tau^w, \tau^l)\}\), the loss function of DPO can be derived:

\vspace{-15pt}
\begin{equation}
\small
\begin{split}
    \mathcal{L}=&-\mathbb{E}_{\left(\tau_w, \tau_l\right) \sim \mathcal{D}}\biggl[\log \sigma\biggl(\left(\sum_{t=0}^{T-1} \beta \log \frac{\pi^*\left(\mathbf{a}_t^w \mid \mathbf{s}_t^w\right)}{\pi_{\mathrm{ref}}\left(\mathbf{a}_t^w \mid \mathbf{s}_t^w\right)}\right) \\
    &-\left(\sum_{t=0}^{T-1} \beta \log \frac{\pi^*\left(\mathbf{a}_t^l \mid \mathbf{s}_t^l\right)}{\pi_{\mathrm{ref}}\left(\mathbf{a}_t^l \mid \mathbf{s}_t^l\right)}\right)\biggr)\biggr]
\end{split}
\end{equation}

\vspace{-5pt}
The additional assumptions of DPO introduce two challenges for multi-step reasoning problems: (1) \textbf{Unnecessary pairwise data collection}: While the BT model is reasonable for a general dialogue system where the reward can only be implicitly inferred from human preference, it's unnecessary for multi-step reasoning tasks where a ground-truth reward exists. To apply DPO on these tasks, previous work has to collect pairwise data on reasoning tasks~\citep{song2024trial, yuan2024advancing}, which is an inefficient usage of offline data. (2) \textbf{No credit assignment}: Relaxing the soft Bellman Equation from every time step to the entire trajectory loses the granularity of credit assignment, which is especially critical in multi-step reasoning tasks where correctness often depends on a few key tokens.

\section{\ours: Offline Reasoning Optimization}
Based on the theorems presented in Section~\ref{sec:prelim}, we present the detailed formulation of our method \ours. We further introduce two objective function variants, an iterative extension of our approach, and a test-time search strategy leveraging the value function.
\label{sec:method}
\subsection{Learning Objetive}
We adopt a similar method as PCL~\citep{nachum2017bridging} to fine-tune the LLM. Inspired by Theorem~\ref{thm:pcl}, we optimize the policy by enforcing the soft Bellman Equation property given in Eq.~\ref{eq:consistency}. In our setting where the reward signal is sparse, we aim to enforce the telescoped version of Eq.~\ref{eq:consistency}, namely
\begin{equation}\label{eq:telescope}
\small
    V^*(\mathbf s_t)=R_t-\beta\sum_{i\ge t}\log\frac{\pi^*(\mathbf a_i|\mathbf s_i)}{\pi_\text{ref}(\mathbf a_i|\mathbf s_i)},
\end{equation}
where $R_t=\sum_{i\ge t} r(\mathbf s_i,\mathbf a_i)$. Note that DPO leverages Eq.~\ref{eq:dpo}, which is a special case of Eq.~\ref{eq:telescope} with $t=0$. We train a separate value network $V_\phi$ together with the policy $\pi_\theta$. We adopt the MSE loss for the value network:
\begin{equation}\label{eq:value}
    \small
    \mathcal L_V(\phi)=\frac 1T\sum_{t=0}^{T-1}\left(V_\phi(\mathbf s_t)-R_t+\beta\sum_{i\ge t}\log\frac{\pi_\theta(\mathbf a_i|\mathbf s_i)}{\pi_\text{ref}(\mathbf a_i|\mathbf s_i)}\right)^2.
\end{equation}
The policy objective is given by
\begin{equation}
\small
\begin{split}
    \mathcal L_\pi(\theta)&=\frac 1T\sum_{t=0}^{T-1}\biggl(V_\phi(\mathbf s_t)-R_t+\beta\log\frac{\pi_\theta(\mathbf a_t|\mathbf s_t)}{\pi_\text{ref}(\mathbf a_t|\mathbf s_t)} \\
    &+\operatorname{sg}\left[\beta\sum_{i>t}\log\frac{\pi_\theta(\mathbf a_i|\mathbf s_i)}{\pi_\text{ref}(\mathbf a_i|\mathbf s_i)}\right]\biggr)^2+\alpha\mathcal L_\text{reg}.
\end{split}
\end{equation}
Here $\operatorname{sg}[\cdot]$ denotes the stop gradient operator, which makes each step have the same scale in the gradient. $\mathcal L_\text{reg}=\frac 1T\sum_{t=0}^{T-1}\operatorname{KL}[\pi_\theta(\cdot|\mathbf s_t)\|\pi_\text{ref}(\cdot|\mathbf s_t)]$ is a regularization term that helps stabilize training.

\subsection{Loss Variants}
In addition to our \ours learning objective, we present two variants: step-level \ours and response-level \ours.

In step-level \ours, an action is considered to be an entire reasoning step instead of a single generated token. For example, \emph{In May, Natalia sold 48 / 2 = 24 clips}. In the context of language models, the probability of taking an action $\mathbf a=(t_1t_2\cdots t_k)$ is $\pi(\mathbf a|\mathbf s)=\prod_{i=1}^kp(t_i|st_1t_2\cdots t_{i-1})$, where $t_i$ denotes the $i$th token of the action and $p$ denotes the language model. The step-level \ours objective can thus be modified accordingly. This objective can also be grounded in the token-level MDP.% (see Appendix~\ref{sec:step_mdp}).

Response-level \ours aims to mimic the behavior of DPO. Instead of enforcing the consistency property at each time step, the action objective considers only the initial state, i.e.,
\begin{equation}
\small
\begin{split}
    \mathcal L^\text{resp}_\pi(\theta)&=\biggl(V_\phi(\mathbf s_0)-R_0 +\beta\sum_{i\ge 0}\log\frac{\pi_\theta(\mathbf a_i|\mathbf s_i)}{\pi_\text{ref}(\mathbf a_i|\mathbf s_i)}\biggr)^2+\alpha\mathcal L_\text{reg}.
\end{split}
\end{equation}

\subsection{Iterative \ours}
\label{sec:iter}
Previous works have shown that offline LLM fine-tuning methods can be applied iteratively to improve model performance~\citep{pang2024iterative, song2024trial, xiong2024iterative}. After each iteration, a new dataset is collected using the updated policy model to generate responses or explore the environment, which is used for further training.  %More specifically, let $M_0$ denote the initial SFT model. One can iteratively collect training set $\mathcal D_i$ using model $M_i$ and optimize for a new model $M_{i+1}$. In this paper, we adopt this idea and implement an iterative version of \ours.

\subsection{Test-Time Search with Value Function}
\label{sec:search}
Recently, inference-time scaling~\citep{hao2024llm,snell2024scaling, wu2024inference} has received significant research attention. One notable approach is the use of Process Reward Models (PRM), which evaluate whether a reasoning step is correct. During the inference time, rather than decoding the reasoning chain autoregressively from the policy model, one can conduct a tree search (e.g., beam search) guided by the PRM. Our method provides a value model for free, which estimates the expected future reward and can be directly used to guide beam search.

In fact, previous PRM methods~\citep{wang2024math, wang2024multi, luo2024improve} train their models using Monte Carlo rollouts, which are essentially similar to the objective used for training the value function in our approach (Eq.~\ref{eq:value}). Our principled formulation removes the need for the extensive heuristic designs commonly required in prior works.

%\shibo{to be polished}.
%Recent works adopt PRMs for test-time search to further improve model performance~\citep{snell2024scaling, wu2024inference, brown2024large, yao2024tree, hao2024llm, liu2024don}. \ours jointly train a policy and a value function, making our method naturally suitable for test-time search.

We implement step-level beam search for math reasoning tasks. At each step, we maintain a set of \( B \) candidate partial trajectories. For each candidate, we generate \( B \) potential next reasoning steps. From the resulting \( B^2 \) candidates, we retain the \( B \) with the highest values.

For embodied agent tasks, where environment dynamics are unknown, beam search is not applicable. Instead, we sample \( K \) actions at each step and select the action with the highest value.

% 
\iffalse
After the policy model is trained 

With reward shaping, we can 

\noindent \textbf{Process Reward Model} The value function can be used as a process reward model. \shibo{via reward shaping}

\citet{rafailov2024r} presents the beam search based on the policy likelihood and connects the methods with 
\fi
\section{Experiments}
% I'll put the numbers here to draw the big picture. I'll do the writing later.

In this section, we evaluate our method \ours in math reasoning and embodied agent tasks. We also demonstrate that the value function trained alongside the policy can further improve model performance at test time through step-level beam search or choosing the best-of-K action.

\paragraph{Datasets and Evaluation Metric.} We adopt the GSM8K~\citep{cobbe2021training} and MATH~\citep{hendrycks2021measuring} dataset for the task of math reasoning. GSM8K is a dataset of grade school math problems. It contains 7473 training problems and 1319 test problems. MATH consists of competition-level math problems, with a training set of size 7500 and a test set of size 5000. All problems in these datasets are labeled with step-by-step ground-truth solutions. We use the script from DeepSeekMath\footnotemark to extract the final answer from the solution and evaluate its correctness.
\footnotetext{\url{https://github.com/deepseek-ai/DeepSeek-Math/tree/main/evaluation/eval}}

We adopt ALFWorld~\citep{shridhar2020alfworld} for the task of embodied agent control. ALFWorld provides interactive TextWorld environments for household tasks. Each task is labeled with an expert trajectory. However, these data do not contain any reasoning process. \citet{song2024trial} annotates 3119 ALFWorld training trajectories with rationales for each step, allowing model training with ReAct-style~\citep{yao2022react} prompting. The evaluation set of ALFWorld contains 140 tasks in seen environments and 134 tasks in unseen environments. We evaluate the success rates of agents in completing the tasks within 40 steps. 

\paragraph{Base Models.} For the math reasoning task, we select \texttt{Qwen2.5-Math-1.5B}~\citep{yang2024qwen2} and \texttt{DeepSeekMath-7B-Instruct}~\citep{shao2024deepseekmath} as our base model. For the embodied agent task, we use \texttt{MiniCPM-2B-dpo-bf16}~\citep{hu2024minicpm} as the base model.

\paragraph{Baseline Methods.} In addition to supervised fine-tuning, we compare our method against three other baselines:
\begin{itemize}
    \item Rejection Sampling: The method uses the successful trajectories in the offline dataset to supervise the policy model. Despite its simplicity, rejection sampling proves to be effective in many reasoning tasks. It's also known as STaR~\citep{zelikman2022star}, RAFT~\citep{dong2023raft}, REST~\citep{gulcehre2023reinforced}, REST$^{\text{EM}}$~\citep{singh2023beyond}.
    \item DPO~\citep{rafailov2024direct} uses offline preference data to solve reasoning tasks~\citep{pang2024iterative, song2024trial}.
    \item KTO~\citep{ethayarajh2024kto} is a popular variant of DPO utilizing the Kahneman-Tversky model of human utility and is able to work on unpaired data.
\end{itemize}
% rejection sampling, direct preference optimization (DPO)~\citep{rafailov2024direct}, and Kahneman-Tversky optimization (KTO)~\citep{ethayarajh2024kto}. \whj{Do I need to comment on each of these baselines a bit?} \whj{Maybe give some implementation details here.}

\begin{figure*}[h]
% todo: align
\centering
\begin{minipage}[t]{.58\textwidth}

    \centering
    \begin{tabular}{c|c|c|c|c}
        \toprule
        \multirow{2}{*}{Methods} & \multicolumn{2}{c|}{Qwen 1.5B} & \multicolumn{2}{c}{DeepSeekMath 7B} \\
         & GSM8K & MATH & GSM8K & MATH \\
        \midrule
        SFT\footnotemark & $73.5$ & $47.5$ & $82.9$ & $46.8$ \\
        \midrule
        Rej. Sampling & $74.9$ & $50.3$ & $83.6$ & $47.2$ \\
        DPO & $74.4$ & $49.2$ & $82.4$ & $47.2$ \\
        KTO & $73.4$ & $48.3$ & $82.5$ & $46.9$ \\
        \midrule
        \ours (Ours) & $\mathbf{77.3}$ & $\mathbf{52.5}$ & $\mathbf{85.9}$ & $\mathbf{49.2}$ \\
        \bottomrule
    \end{tabular}
    \captionof{table}{Results on GSM8K and MATH. \ours yields higher accuracies than the baselines across both datasets and model sizes.}
    \vspace{-5pt}
    \label{tab:math_main}
\end{minipage}
\hfill
\begin{minipage}[t]{.38\textwidth}
    \centering
    % \small
    \begin{tabular}{c|c|c}
        \toprule
        Methods & Unseen & Seen \\
        \midrule
        SFT & $67.2$ & $62.9$ \\
        \midrule
        Rej. Sampling & $68.7$ & $79.3$ \\
        DPO & $69.4$ & $64.3$ \\
        \midrule
        \ours (Ours) & $\mathbf{79.1}$ & $\mathbf{80.7}$ \\
        \bottomrule
    \end{tabular}
    \captionof{table}{Success rates in ALFWorld. \ours consistently outperforms all baselines. \iffalse\shibo{this figure is confusing. The blue and orange bars are meant to be compared.. We should assign each method a color, and group them into two settings.}\fi}
    \label{tab:alfworld_main}
\end{minipage}

\end{figure*}
% I re-run the experiments of rejection sampling & DPO on the new cluster.

We leave implementation details in Appendix~\ref{sec:appendix-details}.

\subsection{Main Results}\label{sec:main_results}
We present the experimental results on mathematical reasoning in Table~\ref{tab:math_main}. Consistent with prior research~\citep{yuan2024advancing, pang2024iterative}, we observe that while DPO provides marginal improvements over the SFT checkpoint used for its initialization, simpler methods such as rejection sampling often outperform DPO. In contrast, \ours demonstrates consistent superiority over all baselines across both datasets (GSM8K and MATH). This improvement is also observed universally across models in the Qwen and DeepSeekMath families. Specifically, for Qwen-2.5-Math 1.5B, \ours achieves a 5.2\% relative improvement over SFT on GSM8K and a 10.5\% improvement on MATH. For DeepSeekMath 7B, despite the SFT checkpoint being heavily tuned with 776K samples~\citep{shao2024deepseekmath}, \ours still delivers meaningful improvements, with relative gains of 3.6\% on GSM8K and 5.1\% on MATH. These results highlight the robustness and effectiveness of our approach across different models and datasets.
\footnotetext{The \texttt{DeepSeekMath-7B-Instruct} model is already supervised fine-tuned~\citep{shao2024deepseekmath}. So the SFT results are directly adopted from their paper.}

The experimental results on ALFWorld, an embodied control task, are presented in Table~\ref{tab:alfworld_main}. \ours outperforms all baselines in both settings. Interestingly, rejection sampling performs well in seen environments within ALFWorld. However, its improvement is marginal in unseen settings, whereas \ours achieves a significant 17.7\% relative improvement over the baseline. Compared to SFT which only learns from successful experience, \ours effectively leverages the failed trajectories, which results in more generalizable capabilities.

\begin{figure}[ht]
    \centering
    %\vspace{-1em}
    \includegraphics[width=0.5\linewidth]{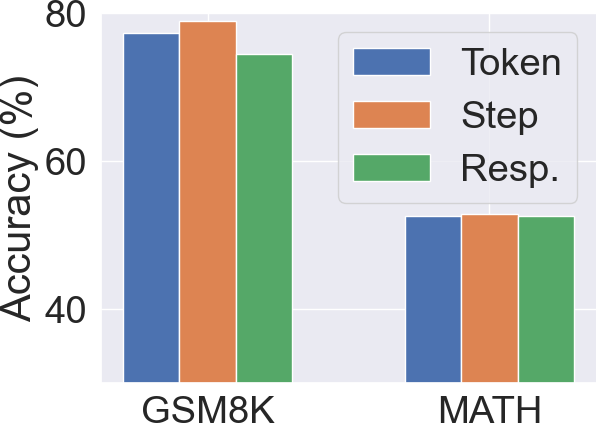}
    \vspace{-5pt}
    \caption{Different variants of \ours objective. ``Token'' stands for the standard \ours objective. ``Step'' denotes the step-level one and ``Resp.'' denotes the response-level one.}
    \label{fig:variant}
    %\vspace{-1em}
    %\vspace{-5pt}
\end{figure}

%\subsection{Different Variants of \ours}
We evaluate different variants of the \ours objective on the math reasoning task. As shown in Table~\ref{fig:variant}, the response-level objective variant performs worse than the token-level objective. This variant treats all actions in the trajectories uniformly, making it challenging to properly assign the sparse reward to individual tokens. This limitation also sheds light on the suboptimal performance of DPO, as it struggles with credit assignment. In contrast, our method explicitly trains a value function, enabling better credit assignment and improved performance. The step-level objective, on the other hand, performs comparably to the token-level objective and even slightly outperforms it on GSM8K. This may be due to the value function’s limited accuracy at each step, which introduces noise into policy learning. Despite the slight performance gap, we adopt the token-level objective as the standard objective in our main experiments due to its simpler implementation (eliminating the need to segment reasoning steps). Nonetheless, step-level policy optimization remains an intriguing avenue for future exploration.

\subsection{Iterative \ours}
\label{sec:iter_exp}
\begin{figure}[ht]
    \centering
    \includegraphics[width=\linewidth]{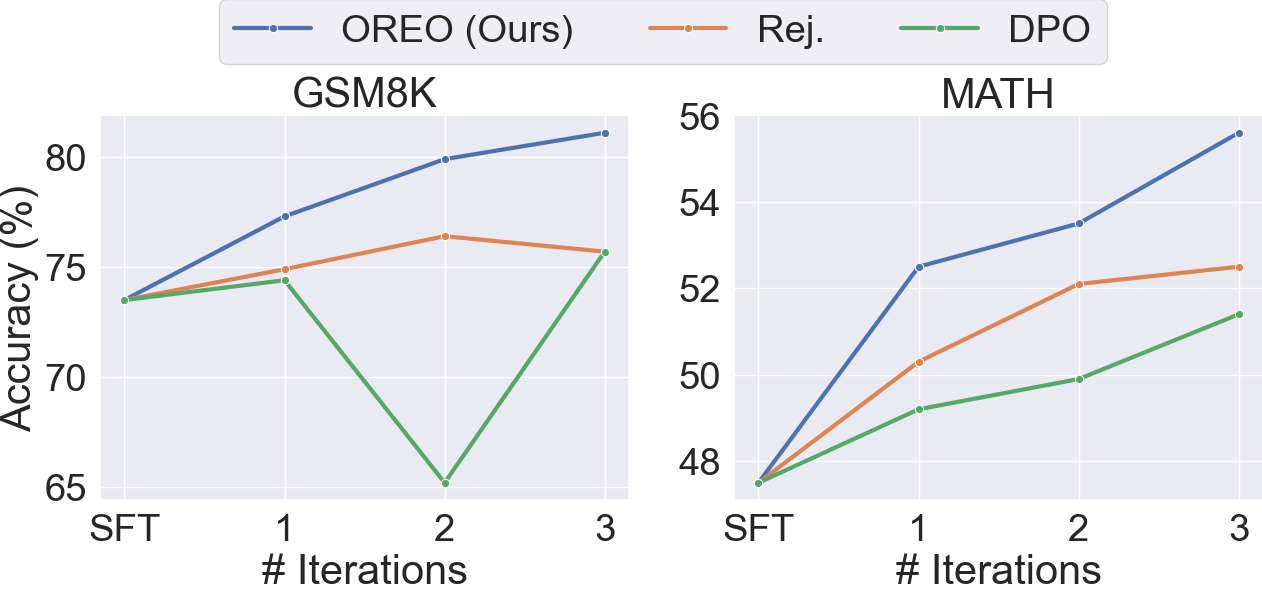}
    \caption{The accuracies on GSM8K and MATH after several iterations. Rej. stands for ``rejection sampling''. \ours improves as new data are collected.}
    \label{fig:iterative}
\end{figure}

Figure~\ref{fig:iterative} illustrates the performance of various algorithms on the math reasoning task across multiple iterations. \ours demonstrates steady and consistent improvements in accuracy over three iterations, showcasing its robustness in leveraging iterative training. While baseline methods also benefit from collecting additional data during each iteration, their performance consistently lags behind that of \ours. Notably, rejection sampling shows signs of saturation by the third iteration, with diminishing performance gains. In contrast, \ours continues to improve, likely due to its ability to effectively learn from failed trajectories. The updated policy model in each new iteration may be able to explore novel failure patterns, and incorporate these insights into the learning process. This potentially explains why \ours benefits more from multiple iterations compared to rejection sampling.

% The performance of the DPO degenerates starting from the second iteration. Rejection sampling also benefits from collecting new data and running multiple iterations. However, it performs worse than \ours in the same number of positive instances.
% todo curve: sft -> 1 iter -> 2 iter -> 3 iter

\begin{figure*}[ht]
    \centering
    \vspace{-1em}
    \includegraphics[width=\linewidth]{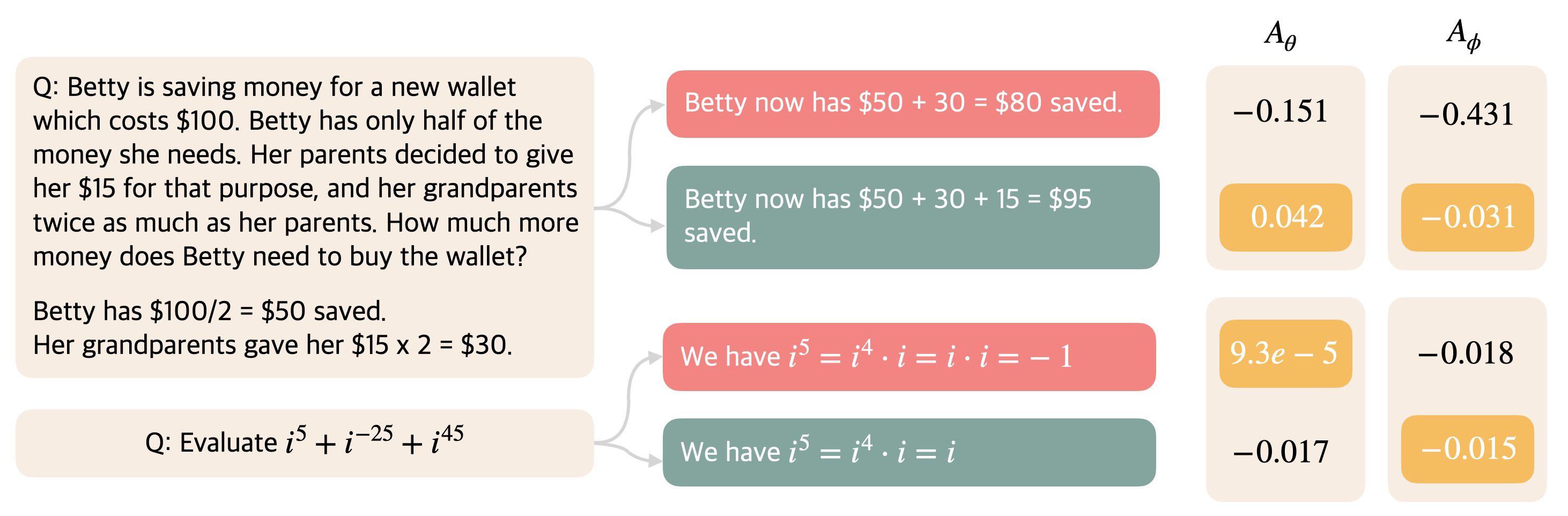}
    \vspace{-2em}
    \caption{Case studies on the implicit and explicit value functions. Correct reasoning steps are shown in green, while incorrect ones are shown in red. Higher advantages predicted by the value functions are highlighted in yellow. Ideally, a good value function should predict a higher advantage for the correct reasoning step.}
    \label{fig:case_study}
    %\vspace{-1em}
\end{figure*}

\begin{figure}[ht]
    \centering
    \includegraphics[width=\linewidth]{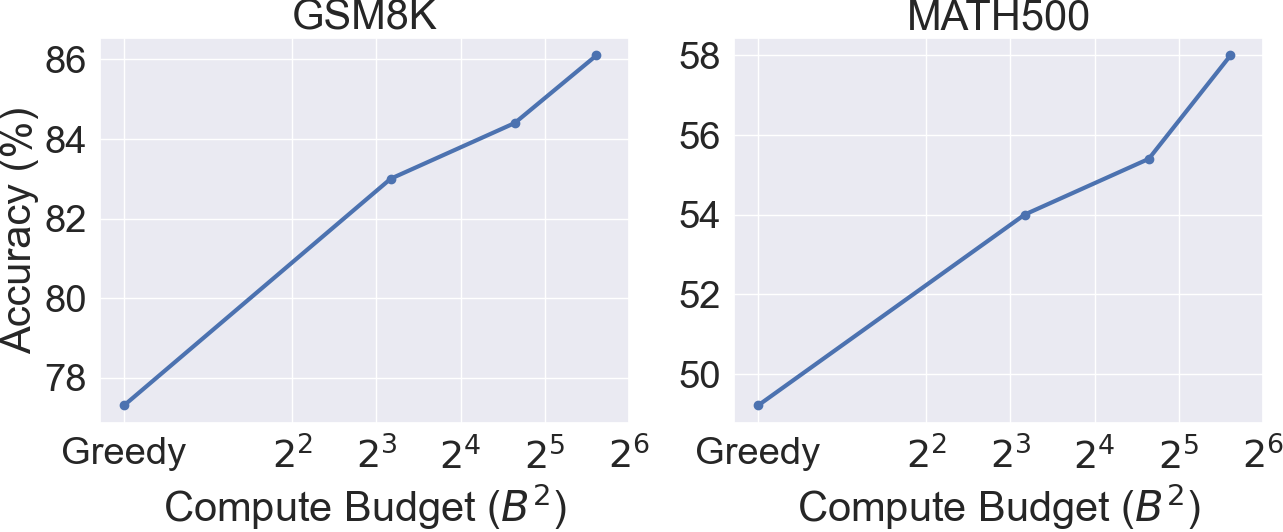}
    \caption{The accuracies of \ours on GSM8K and MATH improve with the compute budget. ``Rej.'' stands for rejection sampling. MATH500 is a subset of MATH containing 500 queries.}
    \label{fig:math_sbs}
\end{figure}

\begin{figure}[ht]
    \centering
    \includegraphics[width=0.5\linewidth]{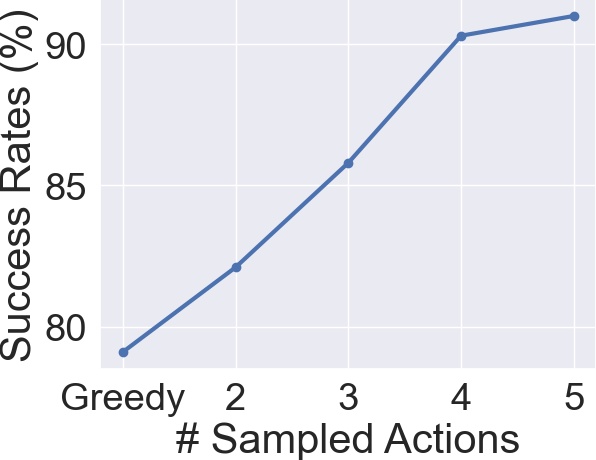}
    %\vspace{-1em}
    \caption{Success rates in ALFWorld when choosing best-of-$K$ actions. Sampling 5 actions and choosing the one with the largest value gains significant improvement in success rates.}
    \label{fig:alfworld_bok}
\end{figure}

% As illustrated in Figure~\whj{todo}, choosing the best-of-5 action gains significant improvement in success rates.

\subsection{Implicit vs Explicit Value Functions}
\label{sec:implicit}
In DPO, the policy model is viewed as an implicit value function~\citep{rafailov2024r}. However, our results in Section~\ref{sec:main_results} have demonstrated that \ours benefits from explicitly parameterizing a separate value function. In this section, we present case studies to compare the explicit value function $V_\phi$ and the implicit value function derived from $\pi_\theta$, aiming to provide an intuitive understanding of their differences.

Our setting is shown in Figure~\ref{fig:case_study}: given a problem and an existing reasoning chain prefix, we evaluate different possible continuations of the next reasoning steps, with different choices of value functions.

Assuming the token indices for the next reasoning step range from \(i\) to \(j\), the advantage function derived from the value model \(V_\phi\) is:

\begin{equation}
\small
    A_\phi = V_\phi(\mathbf{s}_j) - V_\phi(\mathbf{s}_i),
\end{equation}

which quantifies the contribution of the new reasoning step to the expected reward. If the new step introduces an error, the estimated value of the resulting state $s_j$ will be lower than that of the previous state, resulting in a negative advantage.

In contrast, \citet{rafailov2024r} represent an implicit value function using the policy model, defined as:$V_\theta(\mathbf{s}_t)=V(\mathbf{s}_0) + \sum_{t'=0}^{t} \beta \log \frac{\pi_\theta(\mathbf{a}_{t'} \mid \mathbf{s}_{t'})}{\pi_\mathrm{ref}(\mathbf{a}_{t'} \mid \mathbf{s}_{t'})}$. Therefore, the advantage function derived from the policy model \(\pi_\theta\) is:

\vspace{-1em}
\begin{equation}
\small
    A_\theta = V_\phi(\mathbf{s}_j) - V_\phi(\mathbf{s}_i) = \sum_{t=i}^{j-1} \beta \log \frac{\pi_\theta(\mathbf{a}_t \mid \mathbf{s}_t)}{\pi_\mathrm{ref}(\mathbf{a}_t \mid \mathbf{s}_t)}.
\end{equation}

When the soft Bellman equation holds at every timestep, the two advantage functions should be equivalent, which aligns with the assumption of \citet{rafailov2024r}. However, as illustrated in Figure~\ref{fig:case_study}, there are cases where the two advantage functions diverge.

We observed that the magnitude of \(A_\theta\) is generally smaller than that of \(A_\phi\). For example, in the case \textit{“Betty now has \$50 + \$30 = \$80”}, which is incorrect, the value function decreases significantly from \(V_\phi(s_i) = 0.816\) to \(V_\phi(s_j) = 0.385\), resulting in an advantage of \(-0.431\). In contrast, \(A_\theta\) is much more conservative, indicating its weaker ability to distinguish correct and incorrect reasoning steps. While both advantage functions correctly favor the correct response in the GSM8K example above, in a more challenging MATH example below, \(A_\phi\) successfully identifies the correct reasoning step, whereas \(A_\theta\) fails.

These observations are consistent with the effectiveness of searching with an explicit value model~\citep{feng2023alphazero, silver2017mastering, snell2024scaling}. In the context of LLM reasoning, the gap between \(A_\theta\) and \(A_\phi\) may be attributed to the \textit{softmax bottleneck}~\citep{yang2017breaking}. Specifically, predictions for \(\pi_\theta(a_t \mid s_t)\) across different actions \(a_t\) are generated from the same final hidden state, differing only with a linear language model head followed with softmax. In contrast, \(V_\phi(s_{t+1})\) input the entire text sequence to the transformer network, enabling a richer representation. Exploring the gap between policy and value networks presents an intriguing direction for future research.

\subsection{Test-Time Search with Value Functions}
\label{sec:search_exp}
The superiority of the explicit value function motivates its use to enhance inference through search-based methods. For our experiments, we evaluate a subset of the MATH dataset containing 500 queries, a commonly used benchmark in prior work~\citep{lightman2023let, sun2024easy}.

Figure~\ref{fig:math_sbs} shows the performance of step-level beam search in math reasoning. \ours leverages the value function to achieve progressively higher accuracies as the computational budget increases. Compared to greedy decoding, beam search with $B=7$ provides a $11.4\%$ relative improvement in GSM8K and a $17.9\%$ relative improvement in MATH.  This indicates that the explicit value function is more effective than the policy in distinguishing between correct and incorrect reasoning steps.

Similarly, Figure~\ref{fig:alfworld_bok} presents the success rates in ALFWorld when selecting the best-of-K actions. The success rates improve rapidly as the number of sampled actions increases, while stabilizing with five samples. Importantly, the value function is a natural byproduct of the \ours training framework, unlike prior work on PRM, which often involves substantial data engineering and heuristic design efforts.

%This might be due to the L2 loss when optimizing the policy, which encourages a conservative estimation of the value.

% \begin{table}[ht]
%     \centering
%     \begin{tabular}{c|c|c|c}
%         \toprule
%         Value of $B$ & Greedy & 3 & 5 \\
%         \midrule
%         GSM8K & $77.3$ & $83.0$ & $84.4$ \\
%         MATH & $52.5$ & $56.3$ & $58.7$ \\
%         \bottomrule
%     \end{tabular}
%     \caption{Caption}
%     \label{tab:math_sbs}
% \end{table}

% \begin{table}[ht]
%     \centering
%     \begin{tabular}{c|c|c|c|c}
%         \toprule
%         Value of $K$ & Greedy & 3 & 5 & 7 \\
%         \midrule
%         Success Rates & $79.1$ & $85.8$ & $91.0$ & $90.3$ \\
%         \bottomrule
%     \end{tabular}
%     \caption{Success rates on ALFWorld \whj{cite} when choosing best-of-$K$ actions. Sampling 5 actions and choosing the one with the largest value gains significant improvement in success rates.}
%     \label{tab:alfworld_best_of_K}
% \end{table}
\section{Conclusion}
In this paper, we present \underline{O}ffline \underline{RE}asoning \underline{O}ptimization (\ours), an offline RL algorithm for LLM reasoning and embodied LLM agent tasks. \ours leverages the idea of soft Q-learning. It trains an explicit value function together with the LLM policy by optimizing the soft Bellman Equation. This alleviates the need for paired preference data in DPO and enables fine-grained credit assignment among the reasoning steps. In addition, our value function can be used in test-time search to improve the model performance. We evaluate our method in GSM8K, MATH, and ALFWorld, demonstrating a consistent improvement compared to previous offline RLHF methods like DPO.

\section{Limitations}
\iffalse
The performance of \ours still falls behind LLMs trained with online RL methods~\citep{shao2024deepseekmath,yang2024qwen2}. \shibo{do we really have evidence about that? we should write the limitation carefully so that reviewers cannot make them reasons to reject the paper.} We believe this gap may be mitigated by iteratively collecting new training data. The experiments mainly focus on 1.5B models. We plan to run experiments with larger scale in the future.
\fi
Due to limited computation resources, some of our experiments, including ablation studies, iterative \ours, and test-times search, use 1.5B models. We plan to run experiments on larger scales in the future. Our method has primarily been evaluated on mathematical reasoning and embodied agent tasks. As future work, we aim to extend \ours to a wider variety of tasks, such as coding and web browsing, to explore its effectiveness in domains with different structures and requirements. %Our method may be extended to a wider variety of tasks, e.g., coding and web browsing.

% Bibliography entries for the entire Anthology, followed by custom entries
%\bibliography{anthology,custom}
% Custom bibliography entries only
\bibliography{custom}

\appendix

% \section{...}
\label{sec:appendix}

\section{Implementation Details}
\label{sec:appendix-details}
\subsection{Dataset Construction}
% We use the datasets GSM8K and MATH to train the SFT model for math reasoning. Then we sample 10 responses for each query in the datasets using the SFT model to construct the \ours training set. Only trajectories with correct answers receive a reward 1 at the terminal state. We balance the number of positive and negative instances for the 7B model. For the task ALFWorld, we use the annotated data from \citet{song2024trial} to train our SFT model. We perform 5 rollouts for each training task to form the \ours training set. Successful trajectories receive a reward 1 at the end of the trajectory.
We use the datasets GSM8K and MATH to train the SFT model for math reasoning. For the task of math reasoning using 1.5B models, we sample 10 responses for each query in the GSM8K dataset and the MATH dataset. Only trajectories with correct answers receive a reward 1 at the terminal state. For DPO, we pair up the positive and negative instances for each query and sample at most 6 pairs without replacement.

When training 7B models, we apply a different data-collecting strategy to balance the number of positive and negative instances. We sample 16 responses for each query and randomly select at most 4 positive instances and 4 negative instances. We then enforce that the number of positive instances does not exceed that of negative instances. For example, if there are only 3 negative instances in the 16 sampled responses, then we only select 3 positive instances instead of 4. However, we make sure that at least 1 positive instance is selected for the query (if there is any). For DPO, we then sample at most 10 pairs of data without replacement.

For the task ALFWorld, we use the annotated data from \citet{song2024trial} to train our SFT model. We perform 5 rollouts for each task in the training set. Successful trajectories receive a reward 1 at the end of the trajectory. We sample at most 5 preference pairs for DPO.

\subsection{Segmentation of Reasoning Steps}

In step-level beam search and step-level \ours, we use line breaks and periods to indicate the end of a reasoning step. More specifically, we use line breaks in the GSM8K dataset and use both line breaks and periods in the MATH dataset. This is because the GSM8K dataset already split reasoning steps with line breaks.

\subsection{Hyperparameters}
The batch size is set to 128 for all experiments.

\paragraph{SFT.} The \texttt{Qwen2.5-Math-1.5B} model is trained on the GSM8K and MATH dataset for 3 epochs with a learning rate of $2\times 10^{-5}$. The \texttt{DeepSeekMath-7B-Instruct} model is already instruction fine-tuned, so we do not perform any additional SFT. The \texttt{MiniCPM-2B-dpo-bf16} model is trained on the annotated dataset from \citet{song2024trial} for 2 epochs with a learning rate of $2\times 10^{-5}$.

\paragraph{Rejection Sampling.} All experiments are trained for 1 epoch. For the math reasoning task, the learning rate is set to $5\times 10^{-6}$. For the embodied agent task, the learning rate is $2\times 10^{-5}$.

\paragraph{DPO.} All experiments are trained for 1 epoch with a learning rate of $5\times 10^{-7}$ and $\beta$ set to 0.1.

\paragraph{KTO.} The \texttt{Qwen2.5-Math-1.5B} model is trained for 1 epoch with a learning rate of $5\times 10^{-8}$. The \texttt{DeepSeekMath-7B-Instruct} model is trained for 2 epochs such that the total number of samples matches. The learning rate for the 7B model is $10^{-7}$.

\paragraph{\ours.} The learning rate is $5\times 10^{-6}$. $\beta$ is set to $0.03$ and $\alpha$ is set to $0.01$. The \texttt{Qwen2.5-Math-1.5B} is trained for 1 epoch. The \texttt{DeepSeekMath-7B-Instruct} model is trained for 2 epochs such that the total number of samples matches. The \texttt{MiniCPM-2B-dpo-bf16} model is trained for 3 epochs. To save computation, we use LoRA on the critic for 7B models. The LoRA rank and LoRA alpha are both set to 64. The learning rate for the critic is set to $10^{-4}$.

For step-level \ours and response-level \ours, the loss scales are different, which demands different $\alpha$ in the regularization term. We experiment with $\alpha\in\{0.01,0.1,0.3\}$ and choose the best performance parameter. For step-level \ours, $\alpha=0.1$. For response-level \ours, $\alpha=0.3$.
\section{Proof Sketch of Theorem~\ref{thm:consistency}}
\label{sec:proof}

The RLHF objective can be rewritten as
% \begin{align*}
%     &\quad J_\text{RLHF}(\pi) \\
%     &=\mathbb E_{\tau\sim\pi}\biggl[\sum_{t\ge 0}r(\mathbf s_t,\mathbf a_t) \\
%     &\quad -\beta\log\pi(\mathbf a_t\mid\mathbf s_t)+\beta\log\pi_\text{ref}(\mathbf a_t\mid\mathbf s_t)\biggr] \\
%     &=\mathbb E_{\tau\sim\pi}\biggl[\sum_{t\ge 0}r(\mathbf s_t,\mathbf a_t)+\beta\log\pi_\text{ref}(\mathbf a_t\mid\mathbf s_t) \\
%     &\quad +\beta\mathcal H(\pi(\cdot\mid\mathbf s_t)\biggr],
% \end{align*}
\begin{equation}
\begin{split}
    J_\text{RLHF}(\pi)&=\mathbb E_{\tau\sim\pi}\biggl[\sum_{t=0}^{T-1}r_\text{task}(\mathbf s_t,\mathbf a_t) \\
    &+\beta\log\pi_\text{ref}(\mathbf a_t|\mathbf s_t) +\beta\mathcal H(\pi(\cdot|\mathbf s_t))\biggr].
\end{split}
\end{equation}
So it can be viewed as maximum-entropy RL~\citep{haarnoja2017reinforcement} with reward $r_\text{task}(\mathbf s_t,\mathbf a_t)+\beta\log\pi_\text{ref}(\mathbf a_t|\mathbf s_t)$. Previous works in maximum-entropy RL~\citep{haarnoja2017reinforcement} show that the optimal policy satisfies
\begin{equation}\label{eq:optimal_pi}
    \beta\log\pi^*(\mathbf a_t|\mathbf s_t)=Q^*(\mathbf s_t,\mathbf a_t)-V^*(\mathbf a_t),
\end{equation}
and that the optimal Q function and optimal value function satisfies
\begin{equation}\label{eq:bellman}
    \small
    Q^*(\mathbf s_t,\mathbf a_t)=r(\mathbf s_t,\mathbf a_t) + \mathbb E_{\mathbf s_{t+1}\sim T(\cdot | \mathbf s_t,\mathbf a_t)}[V^*(\mathbf s_{t+1})].
\end{equation}
In RLHF settings, the discount factor $\gamma$ is normally omitted. We combine Eq.~\ref{eq:optimal_pi} and Eq.~\ref{eq:bellman} and apply the reward $r_\text{task}(\mathbf s_t,\mathbf a_t)+\beta\log\pi_\text{ref}(\mathbf a_t|\mathbf s_t)$. This gives us
\begin{equation}
    \small
    V^*(\mathbf s_t)-V^*(\mathbf s_{t+1}) = r_\text{task}(\mathbf s_t,\mathbf a_t)-\beta\log\frac{\pi^*(\mathbf a_t|\mathbf s_t)}{\pi_\text{ref}(\mathbf a_t|\mathbf s_t)}.
\end{equation}

\section{Comparison to \citet{liu2024enhancing}}
\label{sec:liu}

Related to this work, \citet{liu2024enhancing} proposed DQO (Direct Q-function Optimization) to improve LLM multi-step reasoning. Both work builds onto previous maximum entropy RL algorithms, and the resulting algorithms are similar. To be more precise, a special case of their training objective (without using importance sampling or $\lambda$-Return), is equivalent to a special case of ours (token-level \ours, without using the regularization term or stop gradient).

The main differences can be summarized as follows: (1) Our work formulates the joint training of policy and value models by enforcing the soft Bellman Equation in PCL~\citep{nachum2017bridging}, while \citet{liu2024enhancing} derive their method starting from SAC~\citep{haarnoja2018softactorcriticoffpolicymaximum} and eliminate Q-functions with policy and value models. Notably, our derivation highlights the connection to DPO, shedding light on why it struggles with multi-step reasoning.  (2) We include a KL regularization term to stabilize training, and explore two variants of the method to better understand its properties. Along with our open-sourced code, we believe our work can help the community explore best practices for LLM training. (3) Our work presents comprehensive experiments and analyses, including experiments on embodied agent control, the iterative training setting, value-guided tree search, and a comparison between implicit and explicit value functions.

We present the empirical comparison of two methods here. \citet{liu2024enhancing} used \texttt{Qwen2-7B-Instruct} and \texttt{Gemma-1.1-7B-it} in their experiments. As their work is not yet open-sourced, we reproduce DQO and experiment on two math reasoning datasets. We use \texttt{Qwen2.5-Math-1.5B} as the base model, and use the same training data. We report the results in Table~\ref{tab:dqo}.
\begin{table}[ht]
    \centering
    \begin{tabular}{c|c|c}
        \toprule
        Methods & GSM8K & MATH \\
        \midrule
        \ours (Ours) & $77.3$ & $52.5$ \\
        DQO & $75.1$ & $49.4$ \\
        \bottomrule
    \end{tabular}
    \caption{Comparison between \ours and DQO~\citep{liu2024enhancing} in math reasoning tasks.}
    \label{tab:dqo}
\end{table}
\section{Safeguard Statement}
In this paper, we primarily focus on the math reasoning tasks and embodied agent control tasks in a household simulator, posing no significant ethical or harmful concerns. We recognize that future
research on border applications of multi-step reasoning may pose a risk of misuse, and we recommend careful
consideration of all aspects of safety before it’s applied in the real world.

% This is an appendix. Test mathbb \( \mathbb{R}, \mathbb{Z}, \mathbb{N} \)

\end{document}